\pdfoutput=1

\documentclass[11pt]{article}
\usepackage[dvipsnames]{xcolor}
\usepackage[inkscapearea=page,inkscapelatex=false]{svg}

\usepackage[final]{coling}

\usepackage{times}
\usepackage{latexsym}

\usepackage[T1]{fontenc}

\usepackage[utf8]{inputenc}

\usepackage{microtype}

\usepackage{inconsolata}

\usepackage{graphicx}

\usepackage{multirow}
\usepackage{lscape}
\usepackage{svg}
\usepackage{soul}
\usepackage{caption}

\usepackage{subcaption}
\usepackage{kotex}
\usepackage{tikz,siunitx}
\usepackage{etoolbox}
\usetikzlibrary{shapes.geometric,shapes.symbols}

\newcommand{\tikzsymbol}[2][circle]{\tikz[baseline=-0.5ex]\node[inner
sep=2pt,shape=#1,draw,#2]{};}%

\newcommand{\tikzsquare}[2][green,fill=green]{\tikz[baseline=-0.5ex]\draw[#1] (-0.1,-0.1) rectangle (#2,#2);}

\definecolor{mycolor}{RGB}{255,255,0}  

\newrobustcmd*{\mytriangle}[1]{\tikz{\draw[thick, fill=mycolor] (0,0) -- (0.2cm,0) -- (0.1cm,0.2cm) -- cycle;}}

\newcommand{\mg}[1]{\textcolor{black}{#1}}
\newcommand{\bsl}{\textcolor{black}}

\title{Speech Foundation Models and Crowdsourcing for\\Efficient, High-Quality Data Collection}

\author{
 \textbf{Beomseok Lee\textsuperscript{1,2,3}},
 \textbf{Marco Gaido\textsuperscript{2}},
 \textbf{Ioan Calapodescu\textsuperscript{3}},
 \textbf{Laurent Besacier\textsuperscript{3}},
 \textbf{Matteo Negri\textsuperscript{2}}
\\
\\
 \textsuperscript{1}University of Trento, Italy,
 \textsuperscript{2}Fondazione Bruno Kessler, Italy,
 \textsuperscript{3}NAVER LABS Europe, France
\\
 \small{
   \textbf{Correspondence:} \href{mailto:beomseok.lee@unitn.it}{beomseok.lee@unitn.it}
 }
}

\begin{document}
\maketitle

\definecolor{bubbles}{rgb}{0.91, 1.0, 1.0}
\definecolor{electricblue}{rgb}{0.49, 0.98, 1.0}
\definecolor{lightcyan}{rgb}{0.88, 1.0, 1.0}
\definecolor{classicrose}{rgb}{0.98, 0.8, 0.91}
\definecolor{palepink}{rgb}{0.98, 0.85, 0.87}
\definecolor{pastelpink}{rgb}{1.0, 0.82, 0.86}
\definecolor{piggypink}{rgb}{0.99, 0.87, 0.9}

\begin{abstract}


While crowdsourcing is an established solution for facilitating and scaling the collection of speech data, the involvement of non-experts necessitates protocols to ensure final data quality. To reduce the costs of these essential controls, this paper investigates the use of Speech Foundation Models (SFMs) to automate the validation process, examining for the first time the cost/quality trade-off in data acquisition. Experiments conducted on French, German, and Korean data demonstrate that SFM-based validation 
has the potential to 
 reduce reliance on human validation,
resulting in an estimated cost saving of over 40.0\% without degrading final data quality. These findings open new opportunities for more efficient, cost-effective, and scalable speech data acquisition.
\end{abstract}

\section{Introduction}

As in any data-intensive domain, collecting high-quality datasets is a fundamental and costly prerequisite for the development of speech-processing applications. 
Traditional methods heavily rely on human workforce, whose costs, as data collection scales, are hard to sustain. In the quest for scalable solutions to tackle this problem, crowdsourcing 
emerged as a viable option that also enables the coverage of diverse populations \cite{10.1145/2556420.2558858,Poesio2017}.
%
Due to the variable quality of crowd-sourced data, validation methods 
that discard low-quality contributions are essential
to build reliable datasets \cite{negri-etal-2011-divide,sabou-etal-2014-corpus,7456302}. This need is exacerbated in the collection of speech-text pairs, 
where various factors, such as recording equipment and conditions, can introduce errors and inconsistencies that 
compromise 
data quality \citep{disagreement_1,disagreement_2}.
Crowdsourcing this validation process
adds substantial overhead, further 
inflating 
data collection costs.

Recent
advances in foundation models offer new possibilities to enhance the scalability of the process by speeding it up and reducing its inherent costs.
When dealing with textual data,
Large language models (LLMs)
have been successfully used as proxies for human evaluation in tasks like sentiment analysis, machine translation, and text generation \citep{gpt_crowdsource,zheng2023judging,kim2023prometheus}. Similarly, speech foundation models (SFMs) have shown potential not only for evaluating 
synthetic and non-synthetic speech \citep{speech_eval_1,speech_eval_2,speech_eval_3}
but also for
automatizing complex data filtering \citep{transcript_in_crowd} and validation tasks \citep{crowdsouce_data_validation_for_asr}. However, all previous works in the area
focused on improving automatic speech recognition (ASR) performance  rather than optimizing the 
cost-efficiency of data validation 
and
exploring the relationship between these two objectives, which remain underinvestigated aspects.

To fill this gap, this paper explores the use of 
SFMs
to automatize the validation of crowd-sourced speech data. To this aim,
we investigate the employment of off-the-shelf SFMs
such as Whisper and SeamlessM4T \citep{whisper,seamlessm4t-v2}, along with machine translation (MT) models and 
grapheme-to-phoneme conversion (G2P).
%
Through experiments on French, German, and Korean data, we test the integration of 
SFMs and crowdsourcing
to  reduce  validation costs 
while
preserving final data quality.
Our results show that leveraging 
SFMs yields a 
cost
reduction by over 40\%, while maintaining high data quality, significantly improving the efficiency and scalability of crowd-sourced speech data collection.


\section{Context and motivations}

This work stems from experiments conducted both during and after the creation of a multilingual speech
\mg{corpus, \texttt{Speech-MASSIVE} \citep{speech_massive}, covering 12 languages.}
The corpus was crowdsourced,
recruiting native speakers
who were instructed to read aloud short sentences and record their voices under controlled conditions.\footnote{Detailed data collection guidelines emphasized the importance of accurate and natural reading, proper recording conditions, and full adherence to the corresponding text.}
To ensure data quality,
each recording was
validated by crowdsourced human raters, directed to read the original text, listen to the recording, and label it as valid or invalid.  Invalid recordings underwent a second iteration of this two-step recording-validation 
process, which, to prevent endless cycles, concluded after the second validation regardless of the outcome.
%
As a result,
\bsl{\texttt{Speech-MASSIVE}} comprises $84,262$ ($t$, $r$, 
$l$)
triplets for the 12 languages, where $t$ is the original text, $r$ is the acquired recording, and 
$l$
is the  valid/invalid 
label
assigned to $r$.

With \bsl{\texttt{Speech-MASSIVE}} at hand, the goal of the post-hoc experiments documented in this paper was to assess whether the costs of its creation could have been reduced by automating the validation steps. 
Specifically, the objective was to assess whether, and to what extent, transcripts generated by existing SFMs could be leveraged to validate the quality of human-recorded speech. 
Within this framing, we address 
two key
questions: 
\textbf{(1)} Can the distance between SFM-generated transcripts and the original text serve as a reliable proxy for recording quality? \textbf{(2)} With comparable final data quality, what are the cost savings of replacing human validation of recorded speech with SFM-based validation?

\section{Automated validation methods}
\label{sec:methods}

Starting from the ($t$, $r$, $l$) triplets of \bsl{\texttt{Speech-MASSIVE}},
our validation method considers the similarity between the original text ($t$) and the SFM-generated transcripts ($\hat{t}$) of the acquired recordings ($r$) as a proxy for $l$.
To this end, we explored two policies. 
The first policy is a \textbf{distance-based} method that measures the similarity between $t$ and $\hat{t}$ with two widely used edit-distance metrics---Character Error Rate (CER) and Word Error Rate (WER)---and retains triplets with a distance below a specified threshold.
However, this approach may be affected by SFMs' bias towards clean audio or specific accents, potentially invalidating samples with poor recording conditions or strong, distinctive accents.

Our second policy seeks to mitigate this risk by employing a \textbf{decision tree} trained on multiple features. In addition to
CER and WER scores computed as in the 
distance-based method, these features include Translation Error Rate (TER)  and Phoneme Error Rate (PER)  scores, which are also based on edit-distance. TER is computed on the English translations of 
$t$ and $\hat{t}$,
under the assumption that accurate recordings will yield translations 
that closely match 
those of the original text.
A further 
advantage of 
using translations for both 
$t$ and $\hat{t}$
is the normalization of numbers in the resulting texts. 
PER is computed by converting 
$t$ and $\hat{t}$
into phonemes, based on the assumption that this conversion may act as a normalizer for words (e.g. named entities) that were transcribed differently but have similar or identical pronunciations.

\section{Experimental setting}
\label{exp:setting}

\paragraph{Data}
We experiment with three distant languages—Korean, French, and German—out of the 12 covered in \bsl{\texttt{Speech-MASSIVE}}.
\textbf{Korean} triplets are used for a preliminary analysis (\S\ref{subsec:comparison}) 
aimed at comparing our two automated validation methods and selecting the best one. To this end, $5,007$  ($t$, $r$, $l$) triplets were enriched with ``gold''\footnote{As opposed to the ``silver'' ones ($l$) 
produced by crowdsourced annotators.} quality labels ($l^*$), produced by two expert linguists, native Korean speakers. 
A Cohen's Kappa 
($\kappa$ -- \citealt{cohens_kappa})
of $0.82$ on a subset of $100$ common samples indicates `excellent' agreement \citep{fleiss2013statistical} between the two annotators.
On the 
entire annotated set, the 
$\kappa$
between the original silver annotations and the gold 
labels is unsurprisingly lower ($0.65$), though still within the `fair to good' range.
\textbf{French} and \textbf{German} triplets are used in our final experiment 
(\S\ref{sec:applic}), which 
focuses on analyzing the impact of applying the best identified method to
quantify the cost savings
yielded
by SFM-based validation of 
crowdsourced
speech data.

\paragraph{Speech foundation models}
To generate the transcripts ($\hat{t}$) of the acquired recordings, we considered Whisper-large-v3\footnote{\url{https://huggingface.com/openai/whisper-large-v3}} \citep{whisper} and Seamless-m4t-v2-large\footnote{\url{https://huggingface.com/facebook/seamless-m4t-v2-large}} \citep{seamlessm4t-v2}. To identify the better-performing model, we compared their transcription capabilities using the French, German, and Korean test splits of FLEURS \citep{fleurs}, computing CER and WER. Whisper-large-v3 
exhibited
superior performance as reported in \bsl{Table \ref{tab:speech_model_asr}}, justifying its use in our experiments unless otherwise specified. 
\bsl{As described in $\S$\ref{sec:methods}, to assess TER and PER as metrics for normalizing the SFMs' transcription outputs, the NLLB-200\footnote{\url{https://huggingface.com/facebook/nllb-200-distilled-1.3B}} translation model is employed for translation into English, while a neural G2P model\footnote{\url{https://github.com/lingjzhu/CharsiuG2P}} is used to convert graphemes into phonemes.}

\begin{table}[hbt!]
\small
\resizebox{\columnwidth}{!}{%
\begin{tabular}{cc|c|c|}
\cline{3-4}
\multicolumn{1}{l}{}                         & \multicolumn{1}{l|}{} & Whisper & Seamless-m4t \\ \hline
\multicolumn{1}{|c|}{\multirow{2}{*}{de-DE}} & WER               & 4.22    & 31.24        \\ \cline{2-4} 
\multicolumn{1}{|c|}{}                       & CER               & 1.48    & 8.05         \\ \hline
\multicolumn{1}{|c|}{\multirow{2}{*}{fr-FR}} & WER                   & 5.37    & 16.24        \\ \cline{2-4} 
\multicolumn{1}{|c|}{}                       & CER                   & 1.9     & 5.73         \\ \hline
\multicolumn{1}{|c|}{\multirow{2}{*}{ko-KR}} & WER                   & 13.88   & 26.26        \\ \cline{2-4} 
\multicolumn{1}{|c|}{}                       & CER                   & 5.3     & 11.21        \\ \hline
\end{tabular}%
}
\caption{\bsl{CER (↓) and WER (↓) of Whisper-large-v3 and Seamless-m4t-v2-large on FLEURS test data.}}

\label{tab:speech_model_asr}
\end{table}




\begin{figure}[t!]
    \centerline{\includegraphics[width=1.05\columnwidth,keepaspectratio]{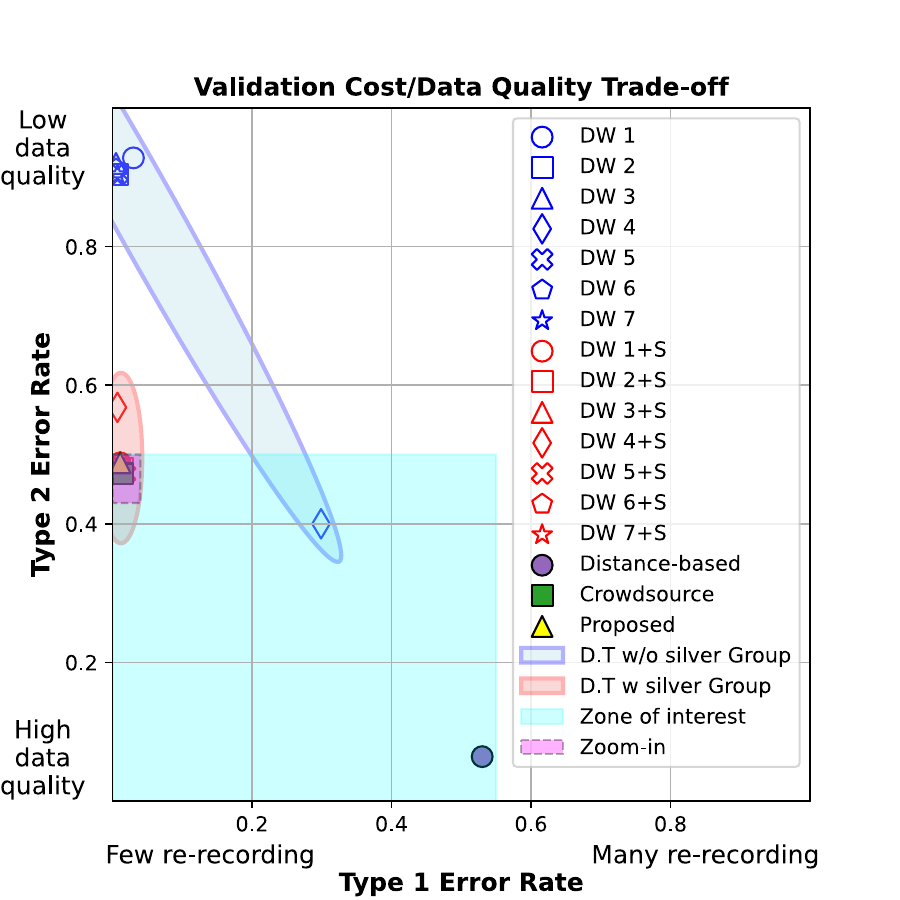}}
    \caption{Results
    of validation methods: 
 DW (decision-tree); DW+S (decision-tree + silver labels); distance-based (simple policy); crowdsource (fully crowdsourced); proposed (final policy for experiments in $\S$\ref{sec:applic})}\label{fig:tradeoff_graph_overview}
\end{figure}

\section{Distance-based method vs Decision Tree}
\label{subsec:comparison}

\begin{figure}[t!]
    \centerline{\includegraphics[width=1.05\columnwidth,keepaspectratio]{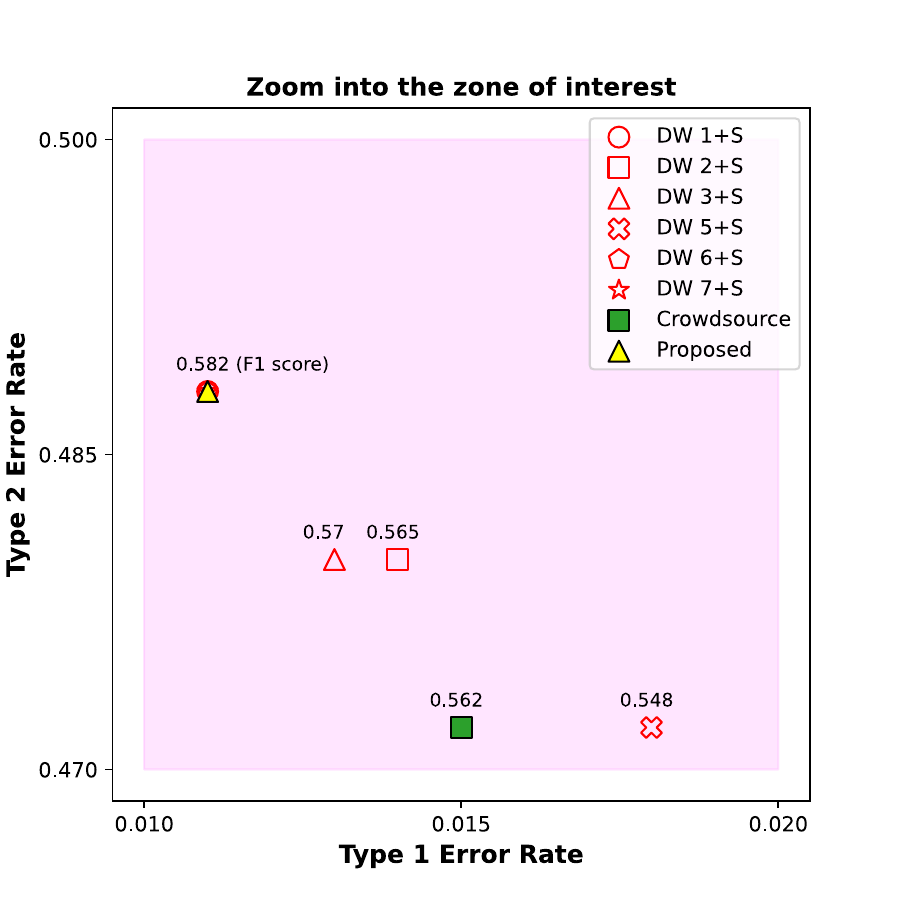}}
    \caption{Zoom-in on a specific area of performance (F1 scores displayed above each data point)}\label{fig:tradeoff_graph_zoom}
\end{figure}

We evaluated various automatic and semi-automatic validation policies by classifying utterances as valid or invalid.
To ensure a fair comparison between the different policies, we use the initial data splits (dev and test) from the \bsl{\texttt{Speech-MASSIVE}} Korean subset. The test split, composed of $2,974$ samples, is used to evaluate and report performance, while the dev split, composed of $2,033$ utterances, is used to train a decision tree when required by the policy. The metrics are calculated by comparing the decisions of the automated validation methods to the gold labels described in \S\ref{exp:setting}. 
We present an overall comparison of the different methods in 
Fig.\ref{fig:tradeoff_graph_overview}, with the exact values  provided  
in Appendix \ref{apdx:metric_comparison}, Table \ref{tab:decision_tree_all}.
Crowdsourced annotations (silver labels) are also evaluated against gold labels to position them on the same evaluation plot as the automated methods.
Specifically, 
Fig.\ref{fig:tradeoff_graph_overview} plots the performance of different methods along \textit{type 1 error rate} (incorrectly classifying valid utterances as invalid, thus requiring re-recording for invalid ones) on the x-axis and \textit{type 2 error rate} (incorrectly classifying invalid utterances as valid) on the y-axis. This visualization highlights the trade-off between the cost of re-recording invalidated utterances (x-axis) and data quality (y-axis). In this context, our 
\colorbox{lightcyan}{zone of interest}
focuses on regions with moderate type 2 error rate (data quality comparable to or better than crowdsourced annotations) and low type 1 error rate (low re-recording costs). 
Fig.\ref{fig:tradeoff_graph_zoom} zooms in on a 
\colorbox{piggypink}{specific area}
of this zone of interest to better distinguish methods 
of similar performance.
%
%
%
%
%
%


\subsection{Distance-based method}
\label{subsec:distance_based_method}

 
The  distance-based method (\tikzsymbol{fill=Periwinkle}) validates recordings only when the ASR output ($\hat{t}$), compared to the reference transcript ($t$), 
shows
both CER and WER values equal to zero. As shown in Fig.\ref{fig:tradeoff_graph_overview}, this policy has high type 1 error rate ($0.53$, meaning that it often invalidates valid samples) 
while at the same time having the lowest type 2 error rate ($0.064$,
crucial for 
high data quality). 
In summary: conservative
and simple (no additional model training is required) \textbf{the distance-based approach prioritizes data quality over reducing re-recording cost}, thus only partially satisfying our requirements.



\subsection{Decision tree}
\label{subsec:decision_tree}

For the decision tree model, we experiment with several settings, testing the two SFMs (Whisper-large-v3 and Seamless-m4t-v2-large) using different 
combinations of CER, WER, PER, and TER features 
($\S$\ref{sec:methods}).
As shown in 
Fig.\ref{fig:tradeoff_graph_overview}, 
basic
decision tree methods ($DW N$, where $N \in [1,7]$, 
full
details in Appendix \ref{apdx:metric_comparison})  
exhibit significantly different behaviour
compared to the distance-based method, 
consistently showing a higher type 2 error rate
across all settings.
%
%
%
Therefore, targeting 
a balance between data quality and cost effectiveness, 
\textbf{basic decision tree models
may not be ideal}, as only one setting ($DW 4$) falls within our zone of interest.
To improve the decision tree model and integrate automation with crowdsourcing,
we incorporate the silver label ($l$) from crowd annotators into our feature set ($DW N + S$).
The red ellipse region in Fig.\ref{fig:tradeoff_graph_overview}, compared to the blue region, shows  the effects of this adjustment in terms of type 2 error rate reduction.
Indeed, most settings of  hybrid (SFMs + crowdsourcing) approach except for $DW 4+S$, fall within our zone of interest.
Examining the zoomed-in area of 
Fig.\ref{fig:tradeoff_graph_zoom}  
reveals
that \textbf{hybrid approaches combining SFMs and crowdsourcing perform comparably to the 
crowdsourcing-only
method} (\tikzsquare[black,fill=Green]{0.15}), suggesting potential cost savings in data validation. 
The next section discusses our final choice of the optimal policy from these options.



\subsection{Proposed method}
Our \textit{proposed} validation method (\mytriangle{mycolor} in Fig.\ref{fig:tradeoff_graph_overview}-\ref{fig:tradeoff_graph_zoom}) 
relies
on a  two-step 
approach.  
First, we use our distance-based method to validate all utterances, minimizing type 2 error rate. For those still flagged as invalid, we use the silver labels assigned by crowd annotators.
Although 
\textbf{the \textit{proposed} approach}
may appear as a simplistic decision tree,
it achieves performance comparable to 
$DW\ 1 + S$ (\tikzsymbol{red} in  in Fig.\ref{fig:tradeoff_graph_overview}-\ref{fig:tradeoff_graph_zoom}) \bsl{as reported in Table \ref{tab:remaining_result}.}
It yields a high F1 score and significantly \textbf{reduces the number of re-recordings required,\footnote{In addition to the re-recording cost, it is important to note that our proposed method incurs costs only for validating the samples flagged as invalid, whereas the $DW N + S$ method requires validation for all samples.}} while maintaining data quality comparable to that achieved through full crowdsourcing.\footnote{We consider the 0.016 difference in type-2 error rates between the two methods to be insignificant, especially given that the manual analysis in Appendix \ref{analysis} highlights disagreements between human (silver and gold) annotations.} In the next section, we 
employ
this automated method for large-scale data validation in German, while using 
French with full crowdsourcing as control language.


\begin{table}[hbt!]
\resizebox{\columnwidth}{!}{%
\begin{tabular}{|l|c|c|c|c|c|}
\hline
 &
  Precision &
  Recall &
  \begin{tabular}[c]{@{}c@{}}F1\\ Score\end{tabular} &
  \begin{tabular}[c]{@{}c@{}}Type 1\\ error\\ rate\end{tabular} &
  \begin{tabular}[c]{@{}c@{}}Type 2\\ error\\ rate\end{tabular} \\ \hline
Distance-based &
  0.072 &
  0.936 &
  0.134 &
  0.530 &
  0.064 \\ \hline
Crowdsource &
  0.600 &
  0.528 &
  0.562 &
  0.015 &
  0.472 \\ \hline
\begin{tabular}[c]{@{}l@{}}Decision Tree \\ ($DW 1 + S$)\end{tabular} &
  0.674 &
  0.512 &
  0.582 &
  0.011 &
  0.488 \\ \hline
Proposed &
  0.674 &
  0.512 &
  0.582 &
  0.011 &
  0.488 \\ \hline
\end{tabular}%
}
\caption{
\bsl{Evaluation results of distance-based (\tikzsymbol{fill=Periwinkle} in Fig.\ref{fig:tradeoff_graph_overview}), crowdsource (\tikzsquare[black,fill=Green]{0.15} in Fig.\ref{fig:tradeoff_graph_overview}-\ref{fig:tradeoff_graph_zoom}), decision tree $DW 1+S$ (\tikzsymbol{red} in  in Fig.\ref{fig:tradeoff_graph_overview}-\ref{fig:tradeoff_graph_zoom})
and proposed method (\mytriangle{mycolor} in Fig.\ref{fig:tradeoff_graph_overview}-\ref{fig:tradeoff_graph_zoom}).}}
\label{tab:remaining_result}
\end{table}

\section{Application to real-world scenarios}
\label{sec:applic}

We conclude by applying our best automated validation method
(\mytriangle{mycolor}) to a real data collection pipeline involving  11,399 new and yet unlabeled samples 
of \bsl{\texttt{Speech-MASSIVE}} German subset. As a term of comparison to assess cost savings of integrating automated validation into the data collection process, we use \bsl{\texttt{Speech-MASSIVE}} French subset, 
entirely (manually) validated through crowdsourcing.


\begin{table}[]
\resizebox{\columnwidth}{!}{%
\begin{tabular}{lccc}
\cline{3-4}
                       & \multicolumn{1}{c|}{}                & \multicolumn{2}{c|}{Cost (\# participants)}                         \\ \cline{3-4} 
                       & \multicolumn{1}{c|}{}                & \multicolumn{1}{c|}{French}      & \multicolumn{1}{c|}{German}      \\ \hline
\multicolumn{1}{|c|}{\multirow{5}{*}{\begin{tabular}[c]{@{}c@{}}1st\\ iteration\end{tabular}}} &
  \multicolumn{1}{c|}{Recording} &
  \multicolumn{1}{c|}{£ 694.73 (572)} &
  \multicolumn{1}{c|}{£ 782.38 (555)} \\ \cline{2-4} 
\multicolumn{1}{|c|}{} &
  \multicolumn{1}{c|}{\begin{tabular}[c]{@{}c@{}}Automated\\ validation\end{tabular}} &
  \multicolumn{1}{c|}{N/A} &
  \multicolumn{1}{c|}{£ 0 (Whisper)} \\ \cline{2-4} 
\multicolumn{1}{|c|}{} & \multicolumn{1}{c|}{\begin{tabular}[c]{@{}c@{}}Human\\ validation\end{tabular}}      & \multicolumn{1}{c|}{£ 333.6 (213)} & \multicolumn{1}{c|}{£ 181.8 (102)} \\ \hline
\multicolumn{1}{|c|}{\multirow{3}{*}{\begin{tabular}[c]{@{}c@{}}2nd\\ iteration\end{tabular}}} &
  \multicolumn{1}{c|}{Recording} &
  \multicolumn{1}{c|}{£ 36 (30)} &
  \multicolumn{1}{c|}{£ 36.4 (26)} \\ \cline{2-4} 
\multicolumn{1}{|c|}{} & \multicolumn{1}{c|}{\begin{tabular}[c]{@{}c@{}}Human\\ validation\end{tabular}}      & \multicolumn{1}{c|}{£ 17.6 (11)}   & \multicolumn{1}{c|}{£ 18 (18)}     \\ \hline
                       & \multicolumn{1}{l}{}                 & \multicolumn{1}{l}{}             & \multicolumn{1}{l}{}             \\ \cline{2-4} 
\multicolumn{1}{l|}{}  & \multicolumn{1}{c|}{All validations} & \multicolumn{1}{c|}{French}      & \multicolumn{1}{c|}{German}      \\ \cline{2-4} 
\multicolumn{1}{l|}{}  & \multicolumn{1}{c|}{cost}            & \multicolumn{1}{c|}{£ 351.2}       & \multicolumn{1}{c|}{£ 199.8}       \\ \cline{2-4} 
\multicolumn{1}{l|}{}  & \multicolumn{1}{c|}{\# participants} & \multicolumn{1}{c|}{224}         & \multicolumn{1}{c|}{120}         \\ \cline{2-4} 
\end{tabular}%
}
\caption{Automated validation applied to German data, with French as a control for evaluating  cost savings (in parentheses, the number of crowdsourced workers).}
\label{tab:real-world}
\end{table}


Table \ref{tab:real-world} shows the total data collection costs for the French and German \bsl{\texttt{Speech-MASSIVE}}.
For the German dataset, the SFM validates a large number of utterances with no labor costs.
We observe a 43.11\% cost reduction in the validation phase, leading to substantial savings in both cost and time by minimizing the need for recruiting and managing human raters.
To ensure comparable  quality between German and French \bsl{\texttt{Speech-MASSIVE}} utterances, 
validated using different policies, we present WER and CER metrics for all recordings in Table \ref{tab:dataset_quality}. 
WER as a proxy for data quality shows that our automated validation for German performs similarly to the fully manual process for French.

\begin{table}[]
\resizebox{\columnwidth}{!}{%
\begin{tabular}{|c|c|c|c|c|}
\hline
                      & langs  & \# samples               & WER   & CER  \\ \hline
\multirow{2}{*}{\bsl{\texttt{Speech-MASSIVE}}} & French & \multirow{2}{*}{11,399} & 11.09 & 4.84 \\ \cline{2-2} \cline{4-5} 
                      & German &                         & 11.7  & 4.19 \\ \hline
\end{tabular}%
}
\caption{Final dataset quality (WER, CER) comparisons using Whisper-v3-large.}
\label{tab:dataset_quality}
\end{table}


\section{Conclusion}


We proposed using Speech Foundation Models (SFMs) to reduce the costs of validating speech data collected through crowdsourcing. 
After exploring various approaches under controlled conditions, we identified a two-step method leveraging Whisper-large-v3 as the most promising. Its application to large-scale validation on German data resulted in a 40\% cost reduction without compromising data quality, demonstrating the strong potential of SFMs to enhance the efficiency and scalability of crowd-sourced speech data collection.

\section{Limitations}
As our proposed method is developed by evaluating only with Korean gold labels, our method lacks language universal development. However, collecting gold annotations for different language for around 5,000 examples is significantly costly. If the proposed methods were developed from various numbers of languages, more universal pattern or method could have been proposed. However, even with this limitation, our proposed method has proven its effectiveness by being successfully applied to German, and further validated through comparison with French to assess dataset quality.

Moreover, this work has limitation with the inherent error in the text corpus which \bsl{\texttt{Speech-MASSIVE}} is built upon. 
As discussed in Appendix \ref{analysis}, some incorrect validations from crowdsource workers are likely due to pre-existing errors in the text. Specifically, problematic prompts affect both the recording and validation phases. During recording, workers are instructed to read the prompt exactly as provided, which can lead to confusion when the prompt is erroneous. Additionally, if workers correct errors in the prompt while recording, it may cause confusion for validators, as the recorded audios are correct despite the original prompt being incorrect.
However, our manual analysis revealed that such cases are relatively rare (6 instances in total—row E in Table \ref{tab:gold_invalid_silver_valid} and row GG in Table \ref{tab:gold_valid_silver_invalid}, out of 2,974 examples) and are unlikely to significantly impact the overall findings.

\section{Acknowledgement}
\bsl{
The research work presented in this paper has been partially funded by the European Union's Horizon research and innovation programme under grant agreement No 101135798, project Meetween (My Personal AI Mediator for Virtual MEETtings BetWEEN People) and the PNRR project FAIR -  Future AI Research (PE00000013),  under the NRRP MUR program funded by the NextGenerationEU.
The crowdsourcing study was funded by the project UTTER (Unified Transcription and Translation for Extended Reality) funded by European Union’s Horizon Europe Research and Innovation programme under grant agreement number 101070631.}






\bibliography{main}

\begin{thebibliography}{21}
\providecommand{\natexlab}[1]{#1}

\bibitem[{Cefkin et~al.(2014)Cefkin, Anya, Dill, Moore, Stucky, and Omokaro}]{10.1145/2556420.2558858}
Melissa Cefkin, Obinna Anya, Steve Dill, Robert Moore, Susan Stucky, and Osariemo Omokaro. 2014.
\newblock \href {https://doi.org/10.1145/2556420.2558858} {Back to the future of organizational work: crowdsourcing and digital work marketplaces}.
\newblock In \emph{Proceedings of the Companion Publication of the 17th ACM Conference on Computer Supported Cooperative Work \& Social Computing}, CSCW Companion '14, page 313–316, New York, NY, USA. Association for Computing Machinery.

\bibitem[{Chittilappilly et~al.(2016)Chittilappilly, Chen, and Amer-Yahia}]{7456302}
Anand~Inasu Chittilappilly, Lei Chen, and Sihem Amer-Yahia. 2016.
\newblock \href {https://doi.org/10.1109/TKDE.2016.2555805} {A survey of general-purpose crowdsourcing techniques}.
\newblock \emph{IEEE Transactions on Knowledge and Data Engineering}, 28(9):2246--2266.

\bibitem[{Cohen(1960)}]{cohens_kappa}
Jacob Cohen. 1960.
\newblock A coefficient of agreement for nominal scales.
\newblock \emph{Educational and psychological measurement}, 20(1):37--46.

\bibitem[{Communication et~al.(2023)Communication, Barrault, Chung, Meglioli, Dale, Dong, Duppenthaler, Duquenne, Ellis, Elsahar, Haaheim, Hoffman, Hwang, Inaguma, Klaiber, Kulikov, Li, Licht, Maillard, Mavlyutov, Rakotoarison, Sadagopan, Ramakrishnan, Tran, Wenzek, Yang, Ye, Evtimov, Fernandez, Gao, Hansanti, Kalbassi, Kallet, Kozhevnikov, Gonzalez, Roman, Touret, Wong, Wood, Yu, Andrews, Balioglu, Chen, Costa-jussà, Elbayad, Gong, Guzmán, Heffernan, Jain, Kao, Lee, Ma, Mourachko, Peloquin, Pino, Popuri, Ropers, Saleem, Schwenk, Sun, Tomasello, Wang, Wang, Wang, and Williamson}]{seamlessm4t-v2}
Seamless Communication, Loïc Barrault, Yu-An Chung, Mariano~Coria Meglioli, David Dale, Ning Dong, Mark Duppenthaler, Paul-Ambroise Duquenne, Brian Ellis, Hady Elsahar, Justin Haaheim, John Hoffman, Min-Jae Hwang, Hirofumi Inaguma, Christopher Klaiber, Ilia Kulikov, Pengwei Li, Daniel Licht, Jean Maillard, Ruslan Mavlyutov, Alice Rakotoarison, Kaushik~Ram Sadagopan, Abinesh Ramakrishnan, Tuan Tran, Guillaume Wenzek, Yilin Yang, Ethan Ye, Ivan Evtimov, Pierre Fernandez, Cynthia Gao, Prangthip Hansanti, Elahe Kalbassi, Amanda Kallet, Artyom Kozhevnikov, Gabriel~Mejia Gonzalez, Robin~San Roman, Christophe Touret, Corinne Wong, Carleigh Wood, Bokai Yu, Pierre Andrews, Can Balioglu, Peng-Jen Chen, Marta~R. Costa-jussà, Maha Elbayad, Hongyu Gong, Francisco Guzmán, Kevin Heffernan, Somya Jain, Justine Kao, Ann Lee, Xutai Ma, Alex Mourachko, Benjamin Peloquin, Juan Pino, Sravya Popuri, Christophe Ropers, Safiyyah Saleem, Holger Schwenk, Anna Sun, Paden Tomasello, Changhan Wang, Jeff Wang, Skyler Wang, and Mary
  Williamson. 2023.
\newblock \href {https://arxiv.org/abs/2312.05187} {Seamless: Multilingual expressive and streaming speech translation}.
\newblock \emph{Preprint}, arXiv:2312.05187.

\bibitem[{Conneau et~al.(2023)Conneau, Ma, Khanuja, Zhang, Axelrod, Dalmia, Riesa, Rivera, and Bapna}]{fleurs}
Alexis Conneau, Min Ma, Simran Khanuja, Yu~Zhang, Vera Axelrod, Siddharth Dalmia, Jason Riesa, Clara Rivera, and Ankur Bapna. 2023.
\newblock Fleurs: Few-shot learning evaluation of universal representations of speech.
\newblock In \emph{2022 IEEE Spoken Language Technology Workshop (SLT)}, pages 798--805. IEEE.

\bibitem[{Fleiss et~al.(2013)Fleiss, Levin, and Paik}]{fleiss2013statistical}
Joseph~L Fleiss, Bruce Levin, and Myunghee~Cho Paik. 2013.
\newblock \emph{Statistical methods for rates and proportions}.
\newblock john wiley \& sons.

\bibitem[{Fu et~al.(2024)Fu, Hung, Tsao, and Wang}]{speech_eval_3}
Szu-Wei Fu, Kuo-Hsuan Hung, Yu~Tsao, and Yu-Chiang~Frank Wang. 2024.
\newblock \href {https://arxiv.org/abs/2402.16321} {Self-supervised speech quality estimation and enhancement using only clean speech}.
\newblock \emph{Preprint}, arXiv:2402.16321.

\bibitem[{He et~al.(2024)He, Huang, Ding, Rohatgi, and Huang}]{gpt_crowdsource}
Zeyu He, Chieh-Yang Huang, Chien-Kuang~Cornelia Ding, Shaurya Rohatgi, and Ting-Hao~Kenneth Huang. 2024.
\newblock \href {https://doi.org/10.1145/3613904.3642834} {If in a crowdsourced data annotation pipeline, a gpt-4}.
\newblock In \emph{Proceedings of the CHI Conference on Human Factors in Computing Systems}, CHI '24, New York, NY, USA. Association for Computing Machinery.

\bibitem[{Kim et~al.(2024)Kim, Shin, Cho, Jang, Longpre, Lee, Yun, Shin, Kim, Thorne, and Seo}]{kim2023prometheus}
Seungone Kim, Jamin Shin, Yejin Cho, Joel Jang, Shayne Longpre, Hwaran Lee, Sangdoo Yun, Seongjin Shin, Sungdong Kim, James Thorne, and Minjoon Seo. 2024.
\newblock \href {https://openreview.net/forum?id=8euJaTveKw} {Prometheus: Inducing evaluation capability in language models}.
\newblock In \emph{Proceedings of the 12th International Conference on Learning Representations}.

\bibitem[{Lee et~al.(2024)Lee, Calapodescu, Gaido, Negri, and Besacier}]{speech_massive}
Beomseok Lee, Ioan Calapodescu, Marco Gaido, Matteo Negri, and Laurent Besacier. 2024.
\newblock Speech-massive: A multilingual speech dataset for slu and beyond.
\newblock In \emph{Proc. Interspeech 2024}, pages 817--821.

\bibitem[{Lee and Glass(2011)}]{transcript_in_crowd}
Chia~Ying Lee and James Glass. 2011.
\newblock \href {https://doi.org/10.21437/Interspeech.2011-761} {{A transcription task for crowdsourcing with automatic quality control}}.
\newblock In \emph{Proc. Interspeech 2011}, pages 3041--3044.

\bibitem[{Maiti et~al.(2023)Maiti, Peng, Saeki, and Watanabe}]{speech_eval_1}
Soumi Maiti, Yifan Peng, Takaaki Saeki, and Shinji Watanabe. 2023.
\newblock \href {https://doi.org/10.1109/ICASSP49357.2023.10095710} {Speechlmscore: Evaluating speech generation using speech language model}.
\newblock In \emph{ICASSP 2023 - 2023 IEEE International Conference on Acoustics, Speech and Signal Processing (ICASSP)}, pages 1--5.

\bibitem[{Marge et~al.(2010)Marge, Banerjee, and Rudnicky}]{disagreement_2}
Matthew Marge, Satanjeev Banerjee, and Alexander~I. Rudnicky. 2010.
\newblock \href {https://doi.org/10.1109/ICASSP.2010.5494979} {Using the amazon mechanical turk for transcription of spoken language}.
\newblock In \emph{2010 IEEE International Conference on Acoustics, Speech and Signal Processing}, pages 5270--5273.

\bibitem[{Negri et~al.(2011)Negri, Bentivogli, Mehdad, Giampiccolo, and Marchetti}]{negri-etal-2011-divide}
Matteo Negri, Luisa Bentivogli, Yashar Mehdad, Danilo Giampiccolo, and Alessandro Marchetti. 2011.
\newblock \href {https://aclanthology.org/D11-1062} {Divide and conquer: Crowdsourcing the creation of cross-lingual textual entailment corpora}.
\newblock In \emph{Proceedings of the 2011 Conference on Empirical Methods in Natural Language Processing}, pages 670--679, Edinburgh, Scotland, UK. Association for Computational Linguistics.

\bibitem[{Novotney and Callison-Burch(2010)}]{disagreement_1}
Scott Novotney and Chris Callison-Burch. 2010.
\newblock \href {https://aclanthology.org/N10-1024} {Cheap, fast and good enough: Automatic speech recognition with non-expert transcription}.
\newblock In \emph{Human Language Technologies: The 2010 Annual Conference of the North {A}merican Chapter of the Association for Computational Linguistics}, pages 207--215, Los Angeles, California. Association for Computational Linguistics.

\bibitem[{Phatthiyaphaibun et~al.(2023)Phatthiyaphaibun, Chaksangchaichot, Rakthammanon, Chuangsuwanich, and Nutanong}]{crowdsouce_data_validation_for_asr}
Wannaphong Phatthiyaphaibun, Chompakorn Chaksangchaichot, Thanawin Rakthammanon, Ekapol Chuangsuwanich, and Sarana Nutanong. 2023.
\newblock \href {https://doi.org/10.21437/Interspeech.2023-389} {{Crowdsourced Data Validation for ASR Training}}.
\newblock In \emph{Proc. INTERSPEECH 2023}, pages 551--555.

\bibitem[{Poesio et~al.(2017)Poesio, Chamberlain, and Kruschwitz}]{Poesio2017}
Massimo Poesio, Jon Chamberlain, and Udo Kruschwitz. 2017.
\newblock \href {https://doi.org/10.1007/978-94-024-0881-2_10} {\emph{Crowdsourcing}}, pages 277--295.
\newblock Springer Netherlands, Dordrecht.

\bibitem[{Radford et~al.(2022)Radford, Kim, Xu, Brockman, McLeavey, and Sutskever}]{whisper}
Alec Radford, Jong~Wook Kim, Tao Xu, Greg Brockman, Christine McLeavey, and Ilya Sutskever. 2022.
\newblock \href {https://arxiv.org/abs/2212.04356} {Robust speech recognition via large-scale weak supervision}.
\newblock \emph{Preprint}, arXiv:2212.04356.

\bibitem[{Ravuri et~al.(2024)Ravuri, Cooper, and Yamagishi}]{speech_eval_2}
Aditya Ravuri, Erica Cooper, and Junichi Yamagishi. 2024.
\newblock \href {https://doi.org/10.1109/ICASSPW62465.2024.10626267} {Uncertainty as a predictor: Leveraging self-supervised learning for zero-shot mos prediction}.
\newblock In \emph{2024 IEEE International Conference on Acoustics, Speech, and Signal Processing Workshops (ICASSPW)}, pages 580--584.

\bibitem[{Sabou et~al.(2014)Sabou, Bontcheva, Derczynski, and Scharl}]{sabou-etal-2014-corpus}
Marta Sabou, Kalina Bontcheva, Leon Derczynski, and Arno Scharl. 2014.
\newblock \href {http://www.lrec-conf.org/proceedings/lrec2014/pdf/497_Paper.pdf} {Corpus annotation through crowdsourcing: Towards best practice guidelines}.
\newblock In \emph{Proceedings of the Ninth International Conference on Language Resources and Evaluation ({LREC}'14)}, pages 859--866, Reykjavik, Iceland. European Language Resources Association (ELRA).

\bibitem[{Zheng et~al.(2023)Zheng, Chiang, Sheng, Zhuang, Wu, Zhuang, Lin, Li, Li, Xing, Zhang, Gonzalez, and Stoica}]{zheng2023judging}
Lianmin Zheng, Wei-Lin Chiang, Ying Sheng, Siyuan Zhuang, Zhanghao Wu, Yonghao Zhuang, Zi~Lin, Zhuohan Li, Dacheng Li, Eric Xing, Hao Zhang, Joseph~E. Gonzalez, and Ion Stoica. 2023.
\newblock \href {https://openreview.net/forum?id=uccHPGDlao} {Judging {LLM}-as-a-judge with {MT}-bench and chatbot arena}.
\newblock In \emph{Proceedings of the 37th Conference on Neural Information Processing Systems (Datasets and Benchmarks Track)}.

\end{thebibliography}

\appendix





\section{Decision Trees}
\subsection{Training and hyper parameters}
To identify the optimal decision tree model, we conducted a 10-fold cross-validation using the F1 score as the evaluation metric. To facilitate interpretation of feature contributions, the tree depth was restricted to 3. Detailed training parameters are provided in the Table \ref{tab:hparam}.

\subsection{Performance comparisons between various decision trees}
\label{apdx:metric_comparison}
For the evaluation results of the decision tree method discussed in \S\ref{subsec:decision_tree},  Table \ref{tab:decision_tree_all} presents the various settings used for the decision trees, including: 1) the choice of SFMs (Whisper, Seamless-m4t, or both) for feature extraction, 2) the specific features selected from the SFMs' outputs, and 3) whether the silver label (crowd-sourced) is included as an additional feature.

\subsection{Decision tree interpretation}
\label{apdx:decision_tree_interprete}
\begin{figure*}
        \centering
        \includegraphics[width=\linewidth]{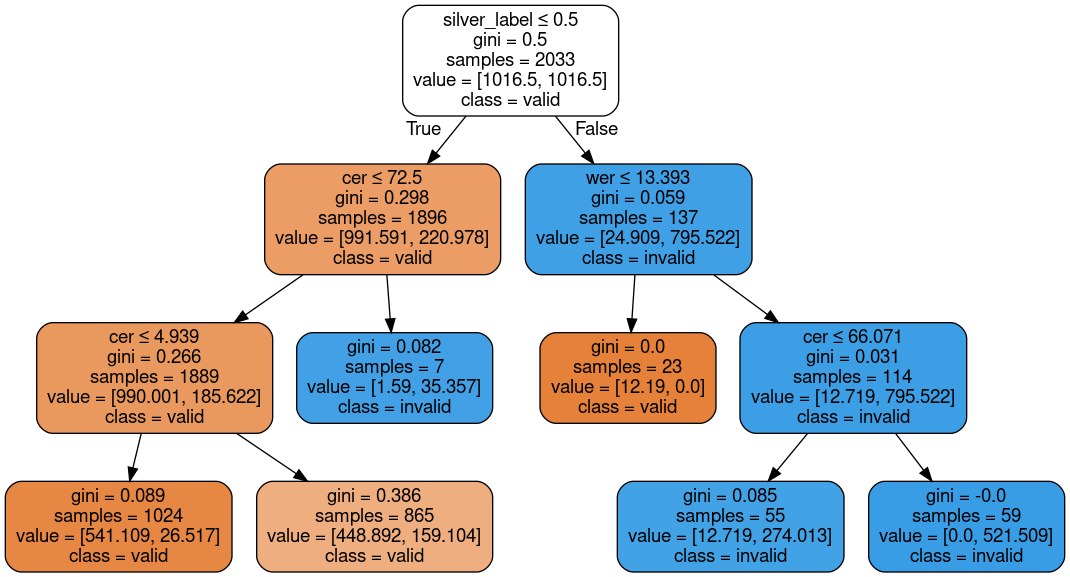}
        \caption{Decision tree graph of DW 5+S method.}\label{fig_apx:decision_tree_DW5+S}
\end{figure*}

We present a decision tree plot to illustrate how features are used to classify samples as valid or invalid. Figure \ref{fig_apx:decision_tree_DW5+S} depicts the decision tree for the $DW\ 5+S$ method, which integrates WER, CER, and silver label features. In this tree, the silver label feature is applied at the root node (level 0) with a threshold of 0.5, meaning that if the silver label equals zero (valid), the samples are further classified by the left child nodes, whereas if the silver label equals $1$ (invalid), the right child nodes continue the classification. 

When the silver label is valid (left child nodes of the root node), the tree uses a CER threshold of $72.5$ at level $1$ to classify certain samples beyond the threshold as invalid, resulting in 7 samples being invalidated (Fig. \ref{fig_apx:decision_tree_DW5+S}). The 1,889 samples where CER is smaller than the CER threshold ($72.5$) are validated.

Conversely, when the silver label is invalid (right child nodes of the root), the tree utilizes a WER threshold of 13.393. If WER is less than or equal to this threshold, 23 samples are validated, otherwise 114 samples are invalidated.

\begin{table}[]
\centering
\begin{tabular}{c|c}
\textbf{parameters} & \textbf{search space} \\ \hline\hline
max depth                      & [1, 2, 3]                                                                          \\ \hline
min samples split              & [3, 5]                                                                        \\ \hline
splitter                       & best, random                                                               \\ \hline
criterion                      & gini, entropy, log\_loss                                                    \\ \hline
class weight                   & balanced, none                                                             \\ \hline
min samples split              & [3, 5, 7, 9, 11, 13]                                                         \\ \hline
min samples leaf               & [1, 3, 5, 7, 9, 11, 13]                                                     
\end{tabular}
\caption{Decision tree fitting hyper parameters.}
\label{tab:hparam}
\end{table}

\section{Confusion matrices of the policies}
\begin{figure*}[]
   \begin{minipage}{0.49\textwidth}
     \centering
     \includegraphics[width=\linewidth]{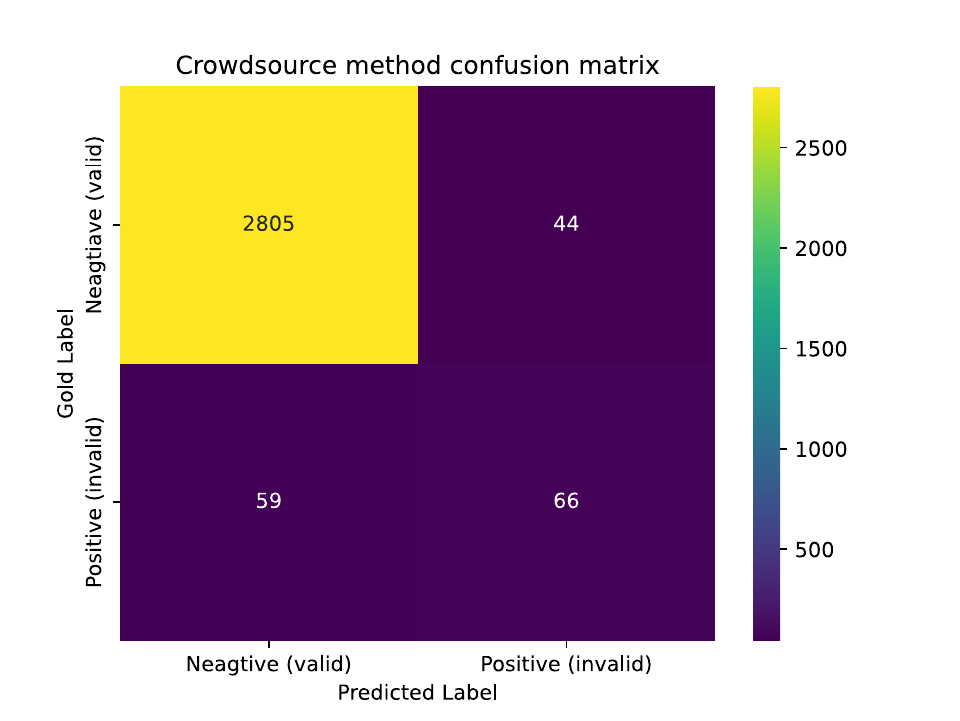}
     \caption{Crowdsource confusion matrix.}\label{fig_apx:confusion_matrix_crowdsource}
   \end{minipage}
   \begin{minipage}{0.49\textwidth}
     \centering
     \includegraphics[width=\linewidth]{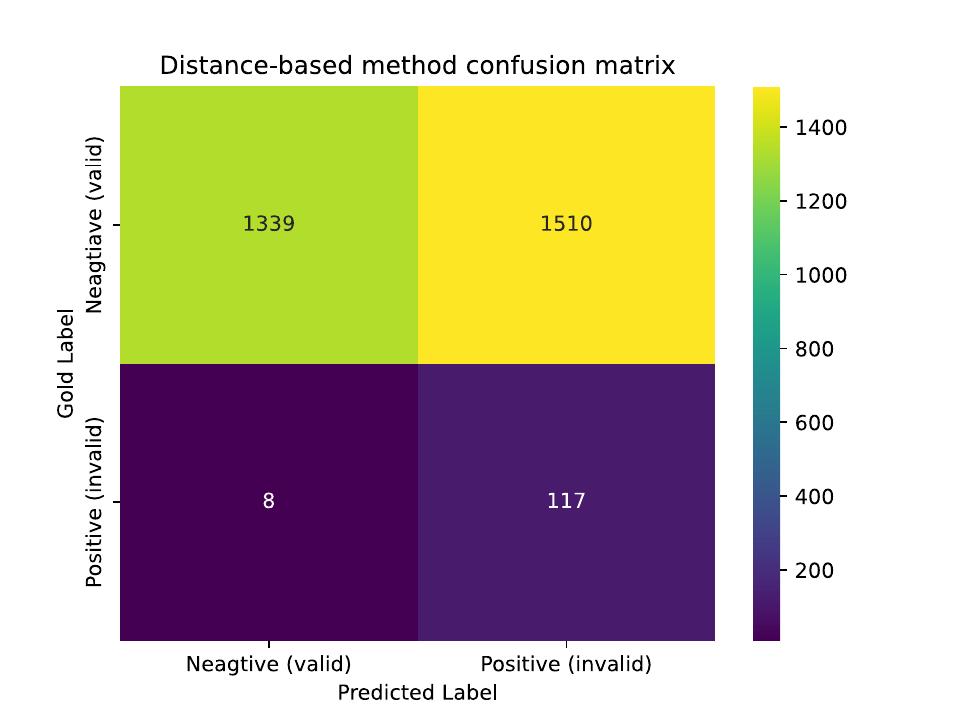}
     \caption{Distance-based method confusion matrix.}\label{fig_apx:confusion_matrix_distance}
   \end{minipage}\hfill
   \begin{minipage}{0.49\textwidth}
     \centering
     \includegraphics[width=\linewidth]{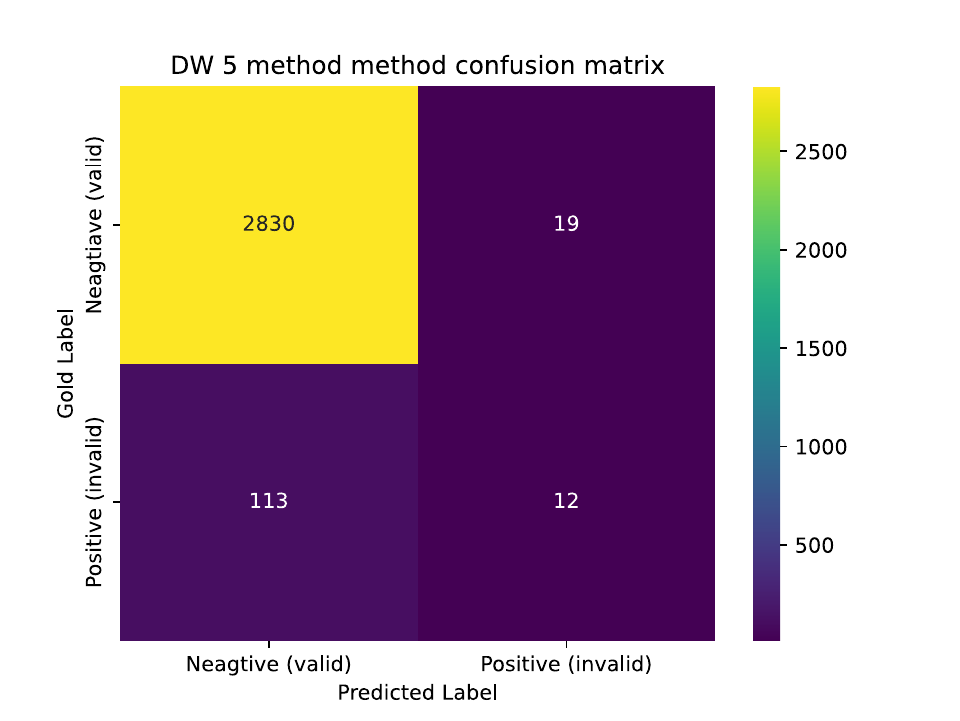}
     \caption{DW 5 method confusion matrix.}
     \label{fig_apx:confusion_matrix_DW5}
   \end{minipage}
   \begin{minipage}{0.49\textwidth}
     \centering
     \includegraphics[width=\linewidth]{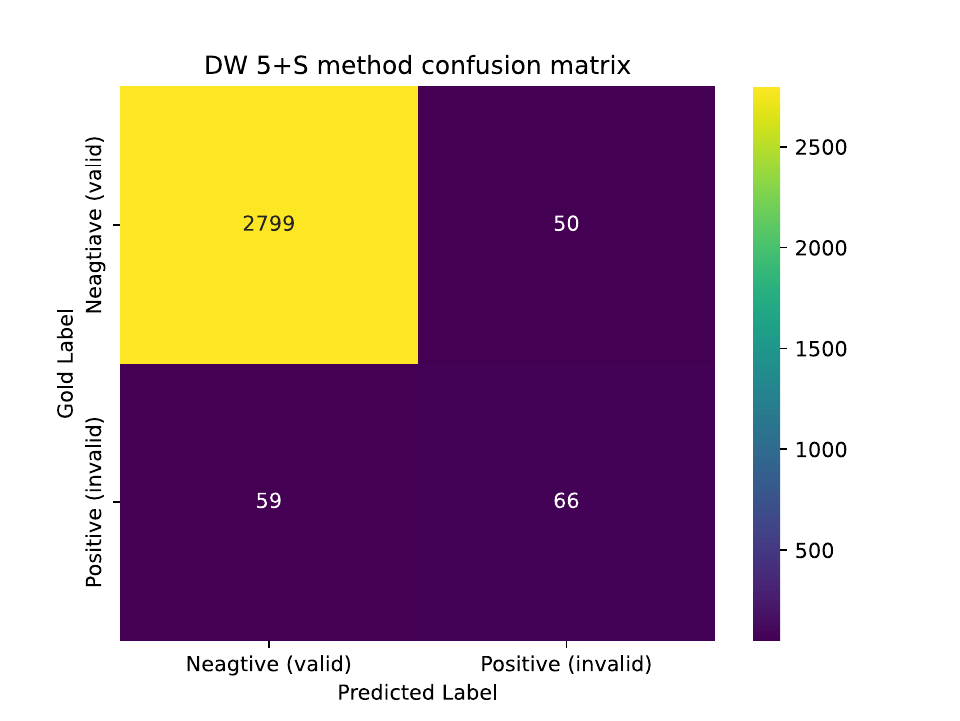}
     \caption{DW 5+S method confusion matrix.}\label{fig_apx:confusion_matrix_DW5+S}
   \end{minipage}
   \begin{minipage}{0.49\textwidth}
     \centering
     \includegraphics[width=\linewidth]{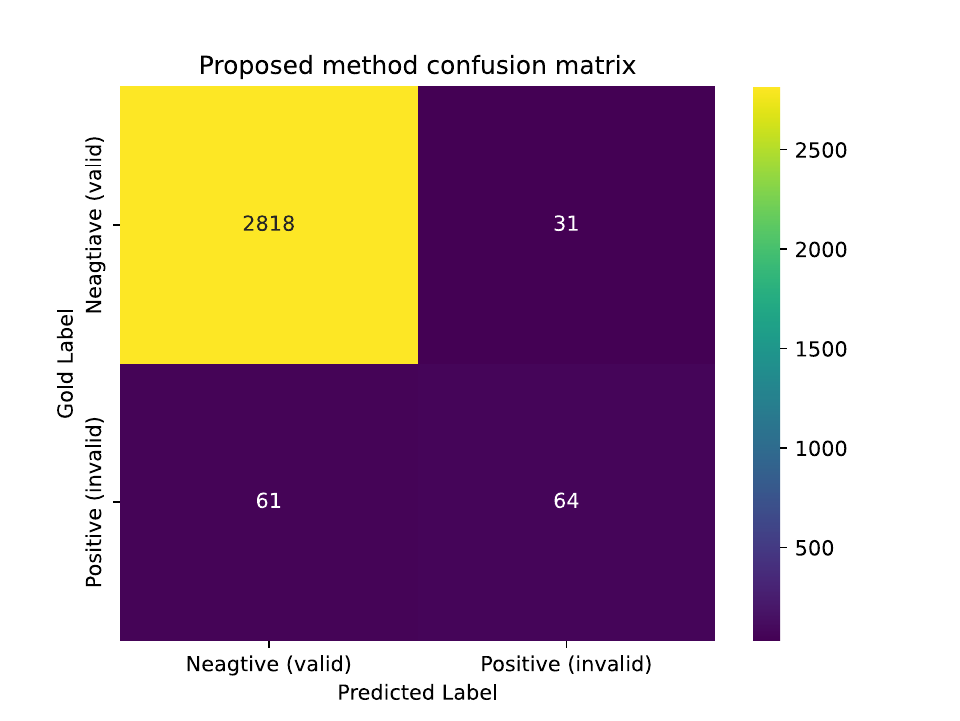}
     \caption{Proposed method confusion matrix.}\label{fig_apx:confusion_matrix_proposed}
   \end{minipage}\hfill
\end{figure*}

\bsl{To supplement the metrics presented in Table \ref{tab:decision_tree_all}, we provide some of the confusion matrix plots in Fig. \ref{fig_apx:confusion_matrix_crowdsource} through Fig. \ref{fig_apx:confusion_matrix_proposed} for various policies and settings as additional material.}

\section{Analysis on gold and silver labels mismatch}
\label{analysis}
In this section, we discuss further analysis on mismatch between gold and silver label for the test split of \bsl{\texttt{Speech-MASSIVE}} Korean subset.

We begin by analyzing annotation mismatches between gold and silver annotations, focusing on cases where the gold annotation is valid, but the silver annotation is invalid. In Table \ref{tab:gold_invalid_silver_valid}. Overall, we witness 47 mismatch cases (44 unique samples) grouped in 10 different types. To further elucidate the types of mismatches observed, consider the following examples. In the $E$ case, the prompt from the corpus is `철수에\textcolor{red}{세} 트윗 남겨', while the correct text should be `철수에\textcolor{blue}{게} 트윗 남겨'\footnote{En Translation: Send a tweet to Cheolsoo.}. The mismatch arises from the incorrect use of `세' instead of the correct `게' resided in the prompt. In the $J$ case, the corpus text `최고로 평점이 좋은 록 음악 팟캐스트 보여줘'\footnote{Show me the best-rated rock music podcast.} contains `록', which is a Koreanization of the English term `rock'. The speaker exhibits `hyperforeignism', pronouncing `록' according to the English pronunciation ($[rOk]$) rather than the Korean phonology ($[lOk]$).

We further classify the mismatches into four categories: \textit{audio}—where the silver annotation's invalidity may be due to poor audio quality, \textit{speaker}—where the invalidity could stem from characteristics of the speaker, \textit{perfect audio}—where the validation is questionable despite clear and intelligible audio, and \textit{corpus}—where the invalidity may result from a typo in the text prompt.
Among the 47 mismatch cases, \textit{audio} category accounts for 16 cases, representing 34\%. \textit{speaker} and \textit{perfect audio} each include 14 cases, which corresponds to 30\% for each category. \textit{corpus} category represents 6\% of the cases.


On the contrary, Table \ref{tab:gold_valid_silver_invalid} presents an analysis of mismatches where the gold is valid, and the silver is invalid, totaling 59 mismatch cases. For $AA$ cases, we observe `near homophone errors'. For instance, `\textcolor{red}{알람}' (alarm, incorrect) is used instead of \textcolor{blue}{`알림'} (notification, correct) in the prompt `내일 오전 열시 미팅 관련해서 리마인더 \textcolor{blue}{알림} 보내줘'\footnote{Send me a reminder notification for the meeting tomorrow morning at 10 AM.}. Such errors often arise when characters or words appear similar in both appearance and sound, leading some crowdsource workers to validate the audio with the homophone error. In $CC$ cases, errors involve the omission, addition, or substitution of particles, possibly due to a lack of attention to grammatical rules or simplification of speech. For example, in the prompt `토요일에 비 소식\textcolor{blue}{이} 있나'\footnote{Is there any news of rain on Saturday?}, one speaker omits the particle `\textcolor{red}{이}', resulting in `토요일에 비 \textcolor{red}{소식} 있나'. For $EE$ cases, approximation errors are noted where speakers slightly modify pronunciation or spelling while maintaining clarity of meaning. For example, in the prompt `지영이한테서 새로 온 \textcolor{blue}{이메일이} 있으면 확인해 줘'\footnote{Check if there is any new email from Jiyoung.}, the speaker says `지영이한테서 새로 온 \textcolor{red}{메일이} 있으면 확인해 줘', substituting `\textcolor{blue}{이메일}' (email) with `\textcolor{red}{메일}' (mail). 

The mismatch types where the gold standard is valid while the silver label is invalid can be further categorized into two groups. \textit{honest mistake} category ($AA\ +\ CC\ +\ EE$) includes cases where validators make errors due to the potentially confusing prompts, accounting for 49\% of the mismatches. \textit{erroneously generous} category ($BB\ +\ DD +\ FF\ + GG$) comprises cases where validators are erroneously generous, representing 51\% of the mismatches.

\definecolor{lightblue}{rgb}{0.68, 0.85, 0.9}
\definecolor{lightred}{rgb}{1.0, 0.7, 0.7}

\begin{table*}[]
\resizebox{\textwidth}{!}{%
\begin{tabular}{cc|l|l|c|c|c|c|c|}
\cline{3-9}
\multicolumn{1}{l}{} &
  \multicolumn{1}{l|}{\textbf{}} &
  \begin{tabular}[c]{@{}c@{}}\textbf{Method}\\ \textbf{name}\end{tabular} &
  \textbf{Features} &
  \textbf{Precision} &
  \textbf{Recall} &
  \textbf{F1 Score} &
  \begin{tabular}[c]{@{}c@{}}\textbf{Type 1}\\ \textbf{error}\\ \textbf{rate}\end{tabular} &
  \begin{tabular}[c]{@{}c@{}}\textbf{Type 2}\\ \textbf{error}\\ \textbf{rate}\end{tabular} \\ \hline
\multicolumn{1}{|c|}{\multirow{11}{*}{without silver label}} &
  \multicolumn{1}{c|}{\multirow{7}{*}{Whisper}} &
  \colorbox{lightblue}{DW 1} &
  WER &
  0.096 &
  0.072 &
  0.082 &
  \colorbox{lightblue}{0.030} &
  \colorbox{lightblue}{0.928} \\ \cline{3-9} 
\multicolumn{1}{|c|}{} & \multicolumn{1}{c|}{}     & \colorbox{lightblue}{DW 2}   & CER           & 0.387 & 0.096 & 0.154 & \colorbox{lightblue}{0.007} & \colorbox{lightblue}{0.904} \\ \cline{3-9} 
\multicolumn{1}{|c|}{} & \multicolumn{1}{c|}{}     & \colorbox{lightblue}{DW 3}   & PER           & 0.435 & 0.080 & 0.135 & \colorbox{lightblue}{0.005} & \colorbox{lightblue}{0.920} \\ \cline{3-9} 
\multicolumn{1}{|c|}{} & \multicolumn{1}{c|}{}     & \colorbox{lightblue}{DW 4}   & TER           & 0.081 & 0.600 & 0.143 & \colorbox{lightblue}{0.299} & \colorbox{lightblue}{0.400} \\ \cline{3-9} 
\multicolumn{1}{|c|}{} & \multicolumn{1}{c|}{}     & \colorbox{lightblue}{DW 5}   & DW 1 + CER    & 0.387 & 0.096 & 0.154 & \colorbox{lightblue}{0.007} & \colorbox{lightblue}{0.904} \\ \cline{3-9} 
\multicolumn{1}{|c|}{} & \multicolumn{1}{c|}{}     & \colorbox{lightblue}{DW 6}   & DW 5 + PER    & 0.440 & 0.088 & 0.147 & \colorbox{lightblue}{0.005} & \colorbox{lightblue}{0.912} \\ \cline{3-9} 
\multicolumn{1}{|c|}{} & \multicolumn{1}{c|}{}     & \colorbox{lightblue}{DW 7}   & DW 6 + TER    & 0.387 & 0.096 & 0.154 & \colorbox{lightblue}{0.007} & \colorbox{lightblue}{0.904} \\ \cline{2-9} 
\multicolumn{1}{|c|}{} &
  \multicolumn{1}{c|}{\multirow{3}{*}{Seamless-m4t}} &
  DS 1 &
  WER &
  0.086 &
  0.336 &
  0.137 &
  0.156 &
  0.664 \\ \cline{3-9} 
\multicolumn{1}{|c|}{} & \multicolumn{1}{c|}{}     & DS 2   & CER           & 0.124 & 0.248 & 0.165 & 0.077 & 0.752 \\ \cline{3-9} 
\multicolumn{1}{|c|}{} & \multicolumn{1}{c|}{}     & DS 3   & DS 1 + CER    & 0.124 & 0.248 & 0.165 & 0.077 & 0.752 \\ \cline{2-9} 
\multicolumn{1}{|c|}{} & \multicolumn{1}{c|}{both} & DWS    & DW 7 + DS 3   & 0.387 & 0.096 & 0.154 & 0.007 & 0.904 \\ \hline
\multicolumn{1}{|c|}{\multirow{11}{*}{with silver label}} &
  \multicolumn{1}{c|}{\multirow{7}{*}{Whisper}} &
  \colorbox{lightred}{DW 1+S} &
  DW 1 + silver &
  0.674 &
  0.512 &
  0.582 &
  \colorbox{lightred}{0.011} &
  \colorbox{lightred}{0.488} \\ \cline{3-9} 
\multicolumn{1}{|c|}{} & \multicolumn{1}{c|}{}     & \colorbox{lightred}{DW 2+S} & DW 2 + silver & 0.619 & 0.520 & 0.565 & \colorbox{lightred}{0.014} & \colorbox{lightred}{0.480} \\ \cline{3-9} 
\multicolumn{1}{|c|}{} & \multicolumn{1}{c|}{}     & \colorbox{lightred}{DW 3+S} & DW 3 + silver & 0.631 & 0.520 & 0.570 & \colorbox{lightred}{0.013} & \colorbox{lightred}{0.480} \\ \cline{3-9} 
\multicolumn{1}{|c|}{} & \multicolumn{1}{c|}{}     & \colorbox{lightred}{DW 4+S} & DW 4 + silver & 0.730 & 0.432 & 0.543 & \colorbox{lightred}{0.007} & \colorbox{lightred}{0.568} \\ \cline{3-9} 
\multicolumn{1}{|c|}{} & \multicolumn{1}{c|}{}     & \colorbox{lightred}{DW 5+S} & DW 5 + silver & 0.569 & 0.528 & 0.548 & \colorbox{lightred}{0.018} & \colorbox{lightred}{0.472} \\ \cline{3-9} 
\multicolumn{1}{|c|}{} & \multicolumn{1}{c|}{}     & \colorbox{lightred}{DW 6+S} & DW 6 + silver & 0.681 & 0.512 & 0.584 & \colorbox{lightred}{0.011} & \colorbox{lightred}{0.488} \\ \cline{3-9} 
\multicolumn{1}{|c|}{} & \multicolumn{1}{c|}{}     & \colorbox{lightred}{DW 7+S} & DW 7 + silver & 0.681 & 0.512 & 0.584 & \colorbox{lightred}{0.011} & \colorbox{lightred}{0.488} \\ \cline{2-9} 
\multicolumn{1}{|c|}{} &
  \multicolumn{1}{c|}{\multirow{3}{*}{Seamless-m4t}} &
  DS 1+S &
  DS 1 + silver &
  0.674 &
  0.480 &
  0.561 &
  0.010 &
  0.520 \\ \cline{3-9} 
\multicolumn{1}{|c|}{} & \multicolumn{1}{c|}{}     & DS 2+S & DS 2 + silver & 0.635 & 0.528 & 0.576 & 0.013 & 0.472 \\ \cline{3-9} 
\multicolumn{1}{|c|}{} & \multicolumn{1}{c|}{}     & DS 3+S & DS 3 + silver & 0.674 & 0.480 & 0.561 & 0.010 & 0.520 \\ \cline{2-9} 
\multicolumn{1}{|c|}{} &
  \multicolumn{1}{c|}{both} &
  DWS+S &
  DW 7 + DS 3 + silver &
  0.681 &
  0.512 &
  0.584 &
  0.011 &
  0.488 \\ \hline
\multicolumn{1}{l}{}   & \multicolumn{1}{l|}{}         & \colorbox{Green}{Crowdsource}    &                      & 0.600 & 0.528 & 0.562 & \colorbox{Green}{0.015} & \colorbox{Green}{0.472} \\ \cline{3-9} 
\multicolumn{1}{l}{}   & \multicolumn{1}{l|}{}         & \colorbox{Thistle}{Distance-based} &                      & 0.072 & 0.936 & 0.134 & \colorbox{Thistle}{0.530} & \colorbox{Thistle}{0.064} \\ \cline{3-9} 
\multicolumn{1}{l}{}   & \multicolumn{1}{l|}{}         & \colorbox{Goldenrod}{Proposed}       &                      & 0.674 & 0.512 & 0.582 & \colorbox{Goldenrod}{0.011} & \colorbox{Goldenrod}{0.488} \\ \cline{3-9} 
\end{tabular}%
}
\caption{Results for all the settings of decision tree and other methods (the color-highlighted settings are the best ones displayed in Fig.\ref{fig:tradeoff_graph_overview}).}
\label{tab:decision_tree_all}
\end{table*}


    
    
    

\begin{table*}[]
\resizebox{2.0\columnwidth}{!}{%
\begin{tabular}{clclc}
\hline
\multicolumn{1}{|c|}{\textbf{type}} &
  \multicolumn{1}{l|}{\textbf{possible reason of silver invalidity}} &
  \multicolumn{1}{c|}{\begin{tabular}[c]{@{}c@{}}\textbf{exclusive}\\ \# \textbf{samples}\end{tabular}} &
  \multicolumn{1}{c|}{\begin{tabular}[c]{@{}c@{}}\textbf{non-exclusive}\\ \# \textbf{samples}\end{tabular}} &
  \multicolumn{1}{c|}{\textbf{category}} \\ \hline
\multicolumn{1}{|c|}{A} & \multicolumn{1}{l|}{perfect audio}             & \multicolumn{1}{c|}{14} & \multicolumn{1}{l|}{}  & \multicolumn{1}{c|}{perfect audio} \\ \hline
\multicolumn{1}{|c|}{B} & \multicolumn{1}{l|}{non-native accent}         & \multicolumn{1}{c|}{7}  & \multicolumn{1}{c|}{1} & \multicolumn{1}{c|}{speaker}       \\ \hline
\multicolumn{1}{|c|}{C} & \multicolumn{1}{l|}{noisy recording condition} & \multicolumn{1}{c|}{5}  & \multicolumn{1}{c|}{2} & \multicolumn{1}{c|}{audio}         \\ \hline
\multicolumn{1}{|c|}{D} & \multicolumn{1}{l|}{rushed speech}             & \multicolumn{1}{c|}{4}  & \multicolumn{1}{c|}{1} & \multicolumn{1}{c|}{speaker}       \\ \hline
\multicolumn{1}{|c|}{E} &
  \multicolumn{1}{l|}{audio-prompt mismatch (*prompt is wrong from the beginning)} &
  \multicolumn{1}{c|}{3} &
  \multicolumn{1}{l|}{} &
  \multicolumn{1}{c|}{corpus} \\ \hline
\multicolumn{1}{|c|}{F} & \multicolumn{1}{l|}{low volume}                & \multicolumn{1}{c|}{3}  & \multicolumn{1}{c|}{2} & \multicolumn{1}{c|}{audio}         \\ \hline
\multicolumn{1}{|c|}{G} &
  \multicolumn{1}{l|}{some strange (mechanical, noise) sound mixed in the audio} &
  \multicolumn{1}{c|}{2} &
  \multicolumn{1}{l|}{} &
  \multicolumn{1}{c|}{audio} \\ \hline
\multicolumn{1}{|c|}{H} &
  \multicolumn{1}{l|}{chopped at the beginning or ending but intelligible} &
  \multicolumn{1}{c|}{1} &
  \multicolumn{1}{l|}{} &
  \multicolumn{1}{c|}{audio} \\ \hline
\multicolumn{1}{|c|}{I} & \multicolumn{1}{l|}{disfluency in the speech}  & \multicolumn{1}{c|}{1}  & \multicolumn{1}{l|}{}  & \multicolumn{1}{c|}{audio}         \\ \hline
\multicolumn{1}{|c|}{J} & \multicolumn{1}{l|}{hyperforeignism}           & \multicolumn{1}{c|}{1}  & \multicolumn{1}{l|}{}  & \multicolumn{1}{c|}{speaker}       \\ \hline
\multicolumn{1}{l|}{}   & \multicolumn{1}{r|}{\# total}                     & \multicolumn{1}{c|}{41} & \multicolumn{1}{c|}{6} &                                    \\ \cline{2-4}
\multicolumn{1}{l|}{}   & \multicolumn{1}{r|}{total \# unique samples}             & \multicolumn{2}{c|}{44}                          &                                    \\ \cline{2-4}
\multicolumn{1}{l}{}    &                                                & \multicolumn{1}{l}{}    &                        &                                   
\end{tabular}%
}
\caption{Analysis on the mismatched annotations between gold (label=invalid) and silver (label=valid). Exclusive samples refer to those that belong to only one group, while non-exclusive samples represent those that are shared among multiple groups.}
\label{tab:gold_invalid_silver_valid}
\end{table*}


\begin{table*}[]
\resizebox{2.0\columnwidth}{!}{%
\begin{tabular}{clccl}
\cline{1-4}
\multicolumn{1}{|c|}{\textbf{type}} &
  \multicolumn{1}{l|}{\textbf{possible reason of silver validity}} &
  \multicolumn{1}{c|}{\textbf{\# samples}} &
  \multicolumn{1}{c|}{\textbf{category}} &
   \\ \cline{1-4}
\multicolumn{1}{|c|}{AA} & \multicolumn{1}{l|}{near homophone error} & \multicolumn{1}{c|}{11} & \multicolumn{1}{c|}{honest mistake}     &  \\ \cline{1-4}
\multicolumn{1}{|c|}{BB} & \multicolumn{1}{l|}{wrong pronunciation}  & \multicolumn{1}{c|}{11} & \multicolumn{1}{c|}{erroneously generous} &  \\ \cline{1-4}
\multicolumn{1}{|c|}{CC} &
  \multicolumn{1}{l|}{particle omission/addition/substitution} &
  \multicolumn{1}{c|}{10} &
  \multicolumn{1}{c|}{honest mistake} &
   \\ \cline{1-4}
\multicolumn{1}{|c|}{DD} &
  \multicolumn{1}{l|}{chopped sentence (at the beginning or ending)} &
  \multicolumn{1}{c|}{9} &
  \multicolumn{1}{c|}{erroneously generous} &
   \\ \cline{1-4}
\multicolumn{1}{|c|}{EE} & \multicolumn{1}{l|}{approximation error}  & \multicolumn{1}{c|}{8}  & \multicolumn{1}{c|}{honest mistake}     &  \\ \cline{1-4}
\multicolumn{1}{|c|}{FF} & \multicolumn{1}{l|}{repetition error}     & \multicolumn{1}{c|}{7}  & \multicolumn{1}{c|}{erroneously generous} &  \\ \cline{1-4}
\multicolumn{1}{|c|}{GG} &
  \multicolumn{1}{l|}{wrong prompt from the beginning and audio not matching the prompt} &
  \multicolumn{1}{c|}{3} &
  \multicolumn{1}{c|}{erroneously generous} &
   \\ \cline{1-4}
\multicolumn{1}{l|}{}    & \multicolumn{1}{r|}{\# total}                & \multicolumn{1}{c|}{59} &                                           &  \\ \cline{2-3}
\multicolumn{1}{l}{}     &                                           &                         &                                           &  \\
\multicolumn{1}{l}{}     &                                           &                         &                                           &  \\
\multicolumn{1}{l}{}     &                                           &                         &                                           &  \\
\multicolumn{1}{l}{}     &                                           &                         &                                           &  \\
\multicolumn{1}{l}{}     &                                           &                         &                                           & 
\end{tabular}%
}
\vspace{-2cm}
\caption{Analysis on the mismatched annotations between gold (label=valid) and silver (label=invalid).}
\label{tab:gold_valid_silver_invalid}
\end{table*}

\end{document}